\documentclass[sigconf,natbib=true,nonacm]{acmart}

\renewcommand\footnotetextcopyrightpermission[1]{}
\setcopyright{none}
\acmConference[AgentSearch 2026]{The First Workshop on Indexing, Retrieval, and Ranking of AI Agents}{July 24, 2026}{Melbourne, Australia}
\acmBooktitle{AgentSearch 2026: The First Workshop on Indexing, Retrieval, and Ranking of AI Agents}
\AtBeginDocument{%
  \pagestyle{fancy}%
  \fancyhf{}%
  \fancyfoot[C]{\thepage}%
}

\usepackage{booktabs}
\usepackage{multirow}
\usepackage{amsmath}
\usepackage{algorithm}
\usepackage{algpseudocode}
\usepackage{graphicx}
\usepackage{xcolor}
\usepackage{subcaption}

\newcommand{\routerec}{\textsc{RouteRec}}

\begin{document}

\title{RouteRec: Strict Evaluation of Recommender-Agent Selection and Aggregation}

\author{Kaiji Zhou}
\affiliation{%
  \institution{University of Birmingham}
  \city{Birmingham}
  \country{United Kingdom}
}
\email{kxz571@student.bham.ac.uk}

\author{Vladimir Kalmykov}
\affiliation{%
  \institution{University of Birmingham}
  \city{Birmingham}
  \country{United Kingdom}
}
\email{vxk276@student.bham.ac.uk}

\author{Yue Feng}
\authornote{Corresponding author.}
\affiliation{%
  \institution{University of Birmingham}
  \city{Birmingham}
  \country{United Kingdom}
}
\email{y.feng.6@bham.ac.uk}

\begin{abstract}
Recommender systems increasingly face a choice among heterogeneous agents---collaborative filters, sequential models, content-based retrievers, and LLM-based rerankers---yet no single agent is uniformly best.
We study this choice as task-aware agent ranking under cost constraints using \routerec{}, a framework that compares request-level hard selection with item-level learned aggregation over four traditional recommender agents and one LLM reranker agent.
On MovieLens-1M, the full quality oracle has substantial headroom (HR@10$=$0.584), confirming that useful cross-agent signal exists.
Under a leakage-free 5-fold out-of-fold protocol, however, hard selection remains below BM25 (0.223 vs.\ 0.254), and selective LLM escalation does not improve it.
The same protocol yields a different outcome for learned aggregation: its cheap-only variant matches BM25 in HR and has a higher NDCG point estimate (0.123 vs.\ 0.114), while gated all-agent aggregation reaches HR@10$=0.295$ with 70.2\% LLM calls.
The resulting lesson is not that routing is solved, but that request-level selection of one complete agent list is too coarse for this sparse fixed-candidate setting; item-level aggregation is the more promising action space.
\end{abstract}

\maketitle

%% ========================================================================
\section{Introduction}
\label{sec:intro}

Recent work has explored LLMs as recommendation agents---as zero-shot rankers~\cite{hou2024llmrank}, instruction followers~\cite{zhang2023instruction}, and tool users~\cite{zhao2024toolrec}.
However, LLM inference can be substantially more expensive than conventional recommendation inference~\cite{chen2023frugalgpt,ong2024routellm}, and LLMs do not uniformly dominate cheaper alternatives.
Traditional recommendation methods remain competitive for warm users with rich interaction histories~\cite{kim2024allmrec}, while surveys discuss potential LLM advantages in sparse, cross-domain, and explanation-oriented settings~\cite{wu2024survey,li2024generative}.

This observation motivates a shift from ``which single model is best?'' to \emph{``which model should handle this request?''}---a framing consistent with classical algorithm selection~\cite{rice1976algorithm}, meta-learning for recommendation~\cite{collins2018aat,cunha2018metalearning,santana2020metabandit}, and LLM routing~\cite{ong2024routellm,shnitzer2023cascading}.

We use \emph{recommender agent} as an operational umbrella term for any callable recommendation service with a ranked-list interface, a cost profile, and task-dependent suitability.
Accordingly, our pool contains four cheap traditional recommender agents and one expensive LLM reranker agent.
This definition matches the agent-search setting, where systems must discover, compare, and invoke heterogeneous callable services; it does not require every service to be an autonomous dialogue agent.

\begin{figure}[t]
\centering
\includegraphics[width=\columnwidth]{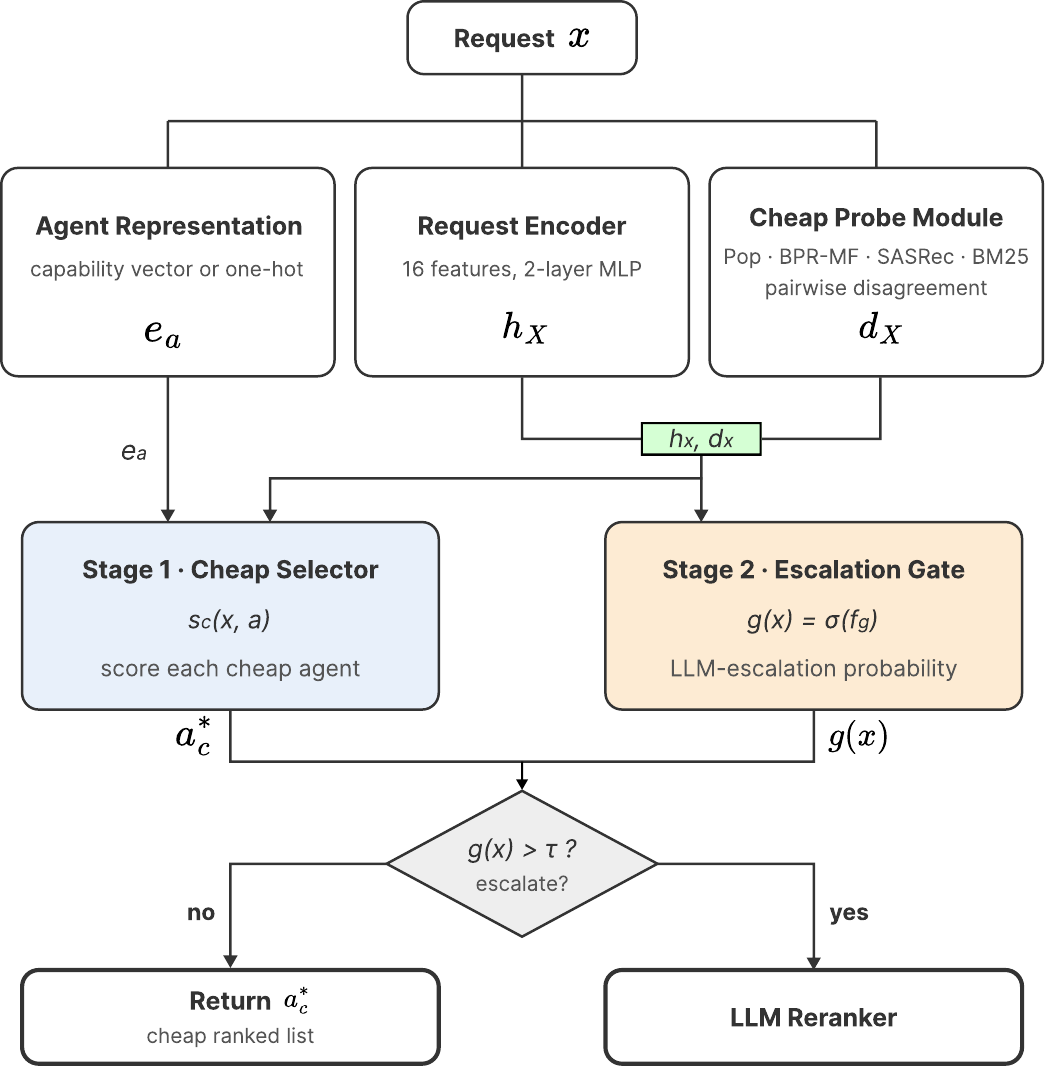}
\caption{\routerec{}-Select architecture. A request is encoded and combined with cheap-probe disagreement features to select the best cheap agent (Stage~1) and decide whether to escalate to the LLM reranker (Stage~2).}
\label{fig:arch}
\Description{System architecture diagram showing the two-stage RouteRec-Select routing pipeline: request encoding, cheap probe disagreement extraction, Stage 1 cheap agent selection, and Stage 2 LLM escalation gate.}
\end{figure}

We study this routing problem through \routerec{}, a lightweight framework for per-instance decisions across the five recommender agents.
\routerec{}-Select combines request context with \emph{cheap probe disagreement}, a deployable measure of cheap-agent output divergence, to choose one cheap agent and decide whether to escalate to the LLM (Fig.~\ref{fig:arch}).
Because hard selection returns one complete list from one agent, it can discard useful evidence from other agents.
We therefore also evaluate \routerec{}-Stack, which reranks the union of agent top-$k$ lists using deployable item-level rank and score evidence.

This paper makes three contributions.
First, we cast search over recommender agents as task-aware utility maximisation under cost constraints and evaluate every trainable policy with 5-fold out-of-fold predictions (Sections~\ref{sec:problem}--\ref{sec:setup}).
Second, we separate deployable signals from oracle labels and show that request-level selection underperforms BM25 despite large oracle headroom (Section~\ref{sec:results}).
Third, we show that learned shortlist aggregation recovers part of the cross-agent signal that hard selection misses, indicating that the unit of the routing decision is a central bottleneck (Section~\ref{sec:discussion}).

%% ========================================================================
\section{Related Work}
\label{sec:related}

\paragraph{LLMs for Recommendation.}
LLM recommendation has moved from prompt reformulation to tuned and tool augmented systems.
Early prompting studies evaluate ChatGPT as a general purpose recommender~\cite{liu2023chatgptrec}.
P5 casts recommendation tasks as language processing~\cite{sun2023p5}.
TALLRec and InstructRec adapt LLMs with supervised instructions~\cite{bao2023tallrec,zhang2023instruction}, while LLM-Rec uses prompting and text enrichment~\cite{lyu2024llmrec}.
LLMRank shows that zero-shot LLMs can rerank candidates, but also exposes position and popularity biases~\cite{hou2024llmrank}.
ToolRec uses LLMs to operate recommender tools~\cite{zhao2024toolrec}.
A-LLMRec connects LLMs with collaborative filtering signals and reports that semantic LLM approaches can be weaker than collaborative methods in warm settings~\cite{kim2024allmrec}.
Surveys similarly emphasise both the promise and the cost, grounding, and evaluation challenges of LLM recommenders~\cite{wu2024survey,li2024generative}.
These findings motivate selective rather than universal LLM deployment.

\paragraph{Hybrid Recommendation and Algorithm Selection.}
Hybrid recommenders have long combined heterogeneous evidence sources such as collaborative, content-based, and knowledge-based signals~\cite{burke2002hybrid}.
The broader algorithm selection problem~\cite{rice1976algorithm} formalises the fact that no single algorithm is universally best.
In recommendation, meta-learning studies select algorithms from observed dataset or request features~\cite{cunha2018metalearning,collins2018aat}, recent work extends this to implicit-feedback ranking tasks~\cite{wegmeth2024algorithm}, and contextual meta-bandits learn online policies over recommender systems~\cite{santana2020metabandit}.
\routerec{} follows this instance-wise selection line, but studies a cost-aware pool containing both traditional recommender agents and an LLM reranker agent under a strict out-of-fold protocol.

\paragraph{LLM Routing and Conditional Computation.}
Mixture-of-Experts models~\cite{jacobs1991moe,shazeer2017moe} provide a general template for conditional computation.
For LLM serving, FrugalGPT~\cite{chen2023frugalgpt}, benchmark-based routing~\cite{shnitzer2023cascading}, and RouteLLM~\cite{ong2024routellm} learn cost-quality tradeoffs between cheaper and stronger LLMs.
LLM-Blender~\cite{jiang2023llmblender} further shows that different LLMs can win on different inputs and that learned pairwise ranking/fusion can exploit this diversity.
\routerec{} differs by routing across \emph{different recommender families} rather than only model sizes, and by relying on cheap recommender probes before any LLM call.

\paragraph{Rank Fusion and Learned Aggregation.}
Simple rank-fusion methods such as reciprocal rank fusion~\cite{cormack2009rrf} combine multiple ranked lists without training, while stacked generalisation~\cite{wolpert1992stacked} and learning-to-rank methods~\cite{liu2009ltr} train second-level scorers over candidate features.
Our learned shortlist aggregation variant applies this idea to the union of recommender top-$k$ lists using only deployable rank, score, membership, and popularity features.
This distinction motivates our empirical split between request-level hard selection and item-level aggregation.

%% ========================================================================
\section{Problem Formulation}
\label{sec:problem}

Using the operational definition above, let $\mathcal{A} = \mathcal{A}_C \cup \mathcal{A}_E$ be a set of recommender agents partitioned into \emph{cheap} ($\mathcal{A}_C$) and \emph{expensive} ($\mathcal{A}_E$) subsets.
For a recommendation request $x$ containing user context, candidate items, and an optional natural-language instruction, each agent $a \in \mathcal{A}$ produces a ranked list $\pi_a(x)$ with accuracy $q(x,a)$ and cost $c(a)$.

Because a deployable policy may run multiple diagnostic probes before returning a final ranked list, utility is defined at the policy level rather than only at the selected-agent level:
\begin{equation}
U(x,\pi,S) = \alpha \cdot \text{NDCG}(x,\pi) + \beta \cdot \text{HR}(x,\pi) - \lambda \cdot \text{lat}(S) - \mu \cdot \text{tok}(S)
\label{eq:utility}
\end{equation}
where $\alpha{=}1, \beta{=}0.5$ weight quality and $\lambda{=}\mu{=}0.01$ penalise latency and token cost respectively.
Here $S$ is the set of agents actually invoked by the policy, $\text{lat}(S)$ is the sum of cached wall-clock execution time in seconds, and $\text{tok}(S)$ is the summed output-token count divided by 1000.
For \routerec{}-Select, $S$ includes the diagnostic probes SASRec, BPR-MF, and BM25, plus the returned agent when it is not already one of those probes; if the gate escalates, $S$ also includes the LLM reranker.
For \routerec{}-StackCheap, $S$ is the cheap-agent list set, while StackAll and StackGate add the LLM only when the all-agent stack is invoked.
The per-instance oracle rows select the agent with the highest quality score ($\alpha\cdot$NDCG$+\beta\cdot$HR), whereas gate and stack thresholds are selected by mean utility on the relevant validation split.
Tables also report HR, NDCG, MRR, and LLM call ratio for interpretability.

The hard-selection goal is to learn a routing policy $\phi(x) \in \mathcal{A}$ that maximises expected policy utility:
\begin{equation}
\max_\phi \, \mathbb{E}_x[U(x, \phi(x), S_\phi(x))]
\end{equation}

\noindent
The implementation decomposes this policy objective rather than predicting exact metrics or calibrated utility.
The cheap selector learns relative scores from quality-only labels conditioned on request, probe, and agent features, while the escalation gate predicts whether the LLM policy has higher cost-aware utility than the best cheap policy.
Section~\ref{sec:stacking} also studies an item-level relaxation that scores candidates from the union of agent shortlists rather than choosing one agent wholesale.

%% ========================================================================
\section{Method}
\label{sec:method}

Fig.~\ref{fig:arch} shows the hard-selection architecture.
\routerec{} has two evaluated forms: \routerec{}-Select, the two-stage request-level router, and \routerec{}-Stack, the shortlist aggregation relaxation.
Both use the same agent pool and deployable shortlist evidence.

\subsection{Agent Pool}
We instantiate $|\mathcal{A}|{=}5$ recommender agents on MovieLens-1M~\cite{harper2015movielens}.
\textbf{Popularity} ranks items by global frequency and serves as the cheapest baseline.
\textbf{BPR-MF}~\cite{rendle2009bpr} is a matrix factorisation model trained with Bayesian Personalized Ranking loss, capturing collaborative signals.
\textbf{SASRec}~\cite{kang2018sasrec} is a self-attentive sequential model for short-term preference dynamics.
\textbf{BM25}~\cite{robertson2009bm25} retrieves over item metadata, using the movie title string (including the original year when present) and genres as document text.
\textbf{LLM Reranker} uses Qwen-2.5-7B-Instruct~\cite{qwen2024} to rerank the first 20 items from a SASRec-seeded candidate pool and is the expensive LLM reranker agent ($\mathcal{A}_E$).
The first four are traditional recommendation agents and form $\mathcal{A}_C$.
As a supplementary comparison, Section~\ref{sec:grok} replaces the Qwen reranker with \texttt{grok-4-1-fast-reasoning}~\cite{xai2026grok41}, a closed-source reasoning model, using the identical prompt and candidate pool construction.
This pool is intentionally diverse, spanning popularity-based, collaborative, sequential, content-based, and language-model-based paradigms.

\subsection{Request Encoder}
\label{sec:request}
Each request $x$ is represented by 16 features across four groups.
The user group contains history length, recency-weighted activity, average inter-event time, category entropy, head-item ratio, and interest drift, measured as the cosine distance between recent and long-term genre distributions.
The candidate group contains pool size, mean popularity, long-tail ratio, and genre diversity of the candidate set.
The task group records whether a natural-language instruction is present, query length, the number of explicit constraints, and comparative wording.
The implementation keeps two system-feature slots for the latency penalty $\lambda$ and token cost penalty $\mu$.
In the present experiments these preferences are fixed for all requests, so after normalisation they provide no meaningful per-request signal; preference-conditioned routing is left for future work.

All continuous features are z-normalised. The concatenated feature vector is passed through a 2-layer MLP ($16 \to 32 \to 32$) with ReLU activations, producing a request embedding $\mathbf{h}_x \in \mathbb{R}^{32}$.

\subsection{Agent Capability Encoder}
\label{sec:capability}
We represent each agent with a 12-dimensional hand-designed \emph{capability vector} $\mathbf{c}_a$.
It covers signal type, instruction support, retrieval/reranking mode, scenario strengths, explanation support, latency, and token cost.

A small MLP ($12 \to 16$) encodes this as $\mathbf{e}_a \in \mathbb{R}^{16}$.
This design enables the router to learn \emph{request--capability} matches rather than memorise per-agent rules~\cite{cunha2018metalearning}.
With a small, closed agent pool ($|\mathcal{A}_C|{=}4$), one-hot encoding performs comparably (Section~\ref{sec:ablation}); whether capability vectors help in larger, open-ended pools remains untested.

\subsection{Cheap Probe Module}
\label{sec:probes}
Before routing, we run three diagnostic cheap probes---SASRec, BPR-MF, and BM25---and compute a 15-dimensional \emph{disagreement vector} $\mathbf{d}_x$ from their outputs.
Popularity remains an eligible cheap agent for final selection, but we exclude it from disagreement probing because its list is mostly global and adds little pairwise diagnostic variation.
\routerec{} uses only deployable probe outputs at inference time; no oracle quality labels are used as input features.
The vector contains four feature families: nine pairwise statistics (top-$k$ overlap, Jaccard similarity, and Kendall $\tau$ for each of the $\binom{3}{2}{=}3$ diagnostic probe pairs), two agreement features (top-1 exact agreement and top-5 set agreement across all probes), two score-level features (mean shortlist-score entropy and mean top-score margin computed from each probe's own returned scores), and two union-list statistics (diversity and mean popularity of the union of all cheap shortlists).
The downstream MLP can weight these derived dimensions unequally, but the module does not impose fixed probe-specific reliability weights.
Because a probe's usefulness may depend on the request, learning explicitly context-dependent probe weights is a question for larger training sets rather than an assumption built into the present model.

\noindent
\textbf{Intuition:} when cheap agents largely agree, the request is likely straightforward and any cheap agent suffices.
When they strongly disagree, there is unresolved ambiguity that may benefit from a more capable (but expensive) agent.
This design is consistent with LLMRank's post-retrieval ranking setup~\cite{hou2024llmrank} and ToolRec's use of LLMs to operate recommender tools~\cite{zhao2024toolrec}.

\subsection{Two-Stage Router}

\paragraph{Stage 1: Cheap Selector.}
For each cheap agent $a \in \mathcal{A}_C$, compute a compatibility score:
\begin{equation}
s_c(x,a) = f_c([\mathbf{h}_x; \mathbf{e}_a; \mathbf{d}_x; p_h(\mathbf{h}_x) \odot \mathbf{e}_a])
\end{equation}
where $p_h(\mathbf{h}_x)=W_h\mathbf{h}_x\in\mathbb{R}^{16}$ projects the 32-dimensional request embedding into the agent-embedding space, $f_c$ is a 2-layer MLP, and $\odot$ denotes element-wise product.
The best cheap agent is $a^*_c = \arg\max_{a \in \mathcal{A}_C} s_c(x,a)$.

\paragraph{Stage 2: Escalation Gate.}
A gating network estimates whether escalation would yield positive marginal utility:
\begin{equation}
g(x) = \sigma(f_g([\mathbf{h}_x; \mathbf{d}_x]))
\end{equation}
If $g(x) > \tau$, the router escalates to the LLM agent; otherwise it uses $a^*_c$.
The final routing policy is:
$$\phi(x) = \begin{cases} a^*_E, & g(x) > \tau \\ a^*_c, & \text{otherwise} \end{cases}$$

This two-stage design makes the key cost decision explicit: (1) choose the best cheap recommendation, then (2) decide whether expensive escalation produces enough marginal utility.

Algorithm~\ref{alg:inference} summarises the \routerec{}-Select inference loop.

\begin{algorithm}[t]
\caption{\routerec{}-Select Inference}
\label{alg:inference}
\begin{algorithmic}[1]
\Require Request $x$, agent pool $\mathcal{A}$, threshold $\tau$
\State Build shared candidate pool $\mathcal{C}(x)$
\State Run diagnostic cheap probes $\to R_1(x), \ldots, R_3(x)$
\State Extract disagreement features $\mathbf{d}_x$ from probe outputs
\State Encode request: $\mathbf{h}_x \gets \psi_x(x)$
\State Select best cheap agent: $a^*_c \gets \arg\max_{a \in \mathcal{A}_C} s_c(x,a)$
\State Estimate escalation gain: $g(x) \gets \sigma(f_g([\mathbf{h}_x; \mathbf{d}_x]))$
\If{$g(x) > \tau$}
    \State \Return LLM agent's ranked list
\Else
    \State \Return $a^*_c$'s ranked list
\EndIf
\end{algorithmic}
\end{algorithm}

\subsection{Training Objective}
\label{sec:training}

The combined loss is $\mathcal{L} = \mathcal{L}_\text{rank} + \eta \cdot \mathcal{L}_\text{esc}$.
The ranking term is a pairwise ranking loss over cheap agents, using \emph{quality-only} labels ($\alpha \cdot$NDCG$+\beta \cdot$HR, \emph{no cost penalty}) to prevent the router from trivially selecting the cheapest agent:
\begin{equation}
\mathcal{L}_\text{rank} = \sum_{(a^+,a^-)} \max(0, m - s_c(x,a^+) + s_c(x,a^-))
\end{equation}
The escalation term $\mathcal{L}_\text{esc}$ is binary cross-entropy for the escalation gate against validation-time labels indicating whether the LLM policy has higher policy-level utility than the best cheap policy under Eq.~\ref{eq:utility}.

\noindent
We use margin $m{=}0.3$, escalation weight $\eta{=}1$, Adam with lr$=10^{-3}$, batch size 64, gradient clipping at 1.0, and early stopping on validation loss with patience 15.
Sample weights are $\max(q_\text{range},0.05)$, where $q_\text{range}$ is the range of cheap-agent quality labels for that request.
We report results averaged over 3~seeds.
The model has \textasciitilde11K parameters and trains in under 3~seconds on CPU.

\subsection{Learned Shortlist Aggregation}
\label{sec:stacking}

Hard agent selection returns one complete list from one agent.
To test whether this action space is too coarse, we evaluate a learned shortlist aggregation variant, also known as stacking.
For each user, the stacker forms a de-duplicated candidate set from the union of agent top-10 lists and assigns a binary label to each item: one for the leave-one-out held-out item and zero for all other union candidates.
No sampled negatives are added beyond this union.
If the held-out item is absent from the union, it is not inserted into the stacker candidate set; that instance has no positive stacker candidate and cannot be hit by any stacking policy.
For each candidate item, the stacker uses only deployable features: per-agent membership, reciprocal rank, rank percentile, within-list min--max normalised score, and train-only item popularity.
The cheap-only stacker selects among balanced logistic-regression models with $C\in\{0.03,0.1,0.3\}$; the all-agent stacker uses the corresponding all-agent union and feature schema and may also select a shallow histogram gradient-boosting classifier.
Model choice is made on the inner validation split by mean utility, with NDCG as a tie-breaker.
We evaluate \emph{StackCheap}, which uses only cheap-agent lists; \emph{StackAll}, which also uses the LLM list; and \emph{StackGate}, which uses cheap-stack confidence to decide whether to invoke StackAll.
The cheap confidence vector contains the top predicted relevance probability, the top1--top2 margin, and score entropy; validation selects whether to threshold top score, margin, or entropy.
This StackGate is distinct from the Stage~2 hard-selection gate: it chooses between cheap-only aggregation and all-agent aggregation, not between one cheap agent and the LLM list.
Budgeted stackers use the same validation procedure under an LLM-ratio cap; the suffix in StackBudget@25, @50, or @75 denotes the validation budget cap, while Table~\ref{tab:main} reports the realised test OOF LLM call ratio.

%% ========================================================================
\section{Experimental Setup}
\label{sec:setup}

\paragraph{Dataset.}
MovieLens-1M~\cite{harper2015movielens} (6,040 users, 3,706 items, 1M ratings).
We use 500 test users from the SASRec evaluation split, with pre-stored LLM reranking logs enabling zero-inference experimentation.
In our offline protocol, each evaluated user corresponds to one recommendation instance; we use ``instance'' and ``request'' interchangeably throughout.
The shared candidate pool is SASRec-seeded and oracle-containing: SASRec first proposes 50 candidates with the held-out target masked, then the held-out target is randomly inserted into an interior position of the cached pool.
All non-LLM agents score and rerank this full shared pool.
The LLM prompt, however, shows only the first 20 candidates under SASRec-seeded order; in the 500-user benchmark the held-out item is visible to the LLM prompt for 36.4\% of requests.
This makes the study a fixed-candidate reranking and routing benchmark, not a full retrieval benchmark, so conclusions about LLM reranking are limited to this candidate-generation protocol.

\paragraph{Agent outputs.}
All agents are run offline and their outputs (ranked lists, scores, latencies, token counts) are stored in JSONL files.
For BM25, each movie document is the title string plus genres.
The query is built only from the synthetic instruction when present and the most recent 20 training-history movies, represented as a title/genre bag; held-out target metadata is never used to form the query.
The LLM agent uses cached Qwen-2.5-7B-Instruct~\cite{qwen2024} logs (temperature 0.1), avoiding re-inference; the prompt shows the first 20 SASRec candidates with titles and genres and asks the model to return a JSON list of candidate positions.
This design eliminates stochastic LLM variation and ensures reproducible evaluation.
Popularity-valued router and stacker features are recomputed from training interactions only; repeating the analysis with stored global popularity counts yields the same qualitative conclusion.

\paragraph{Data split.}
For trainable methods, we use a leakage-free 5-fold out-of-fold protocol over the 500 evaluation users.
Folds are stratified by whether the LLM is utility-best; for each outer fold, the model is trained on 350 users, tuned on 50 validation users, and evaluated on 100 held-out users.
We repeat the full 5-fold procedure for seeds 42, 123, and 456 and report mean$\pm$standard deviation over seeds.
Every reported trainable prediction is therefore made by a model that did not train on that user.
250 of 500 test users receive synthetic natural-language instructions generated from five item-metadata templates over the user's own recent training history, such as genre preferences, negative exclusions, and comparative requests.
The templates use only titles and genres from training-history items and never inspect the held-out target item, preventing target-label leakage through the natural-language request.

\paragraph{Metrics.}
\emph{Quality}: HR@10, NDCG@10, MRR.
\emph{Decision}: utility~$U$, oracle-gap closure, and win rate vs.\ BM25.
\emph{Cost}: LLM call ratio, mean policy latency in seconds, and mean output-token proxy.
\emph{Robustness}: standard deviation over 3~seeds and bootstrap 95\% CI.

\paragraph{Baselines.}
We compare against four categories:
\emph{(i) Single agents}: each of the 5 agents individually;
\emph{(ii) Fusion}: Reciprocal Rank Fusion~\cite{cormack2009rrf}, Borda count, and score averaging, each evaluated both on cheap-only lists and on all lists;
\emph{(iii) Selection}: random agent, single-best-overall, request-feature-only selector, one-hot-agent selector;
\emph{(iv) Quality upper bounds}: instance-level cheap oracle (highest-quality cheap agent per instance) and instance-level full oracle (highest-quality agent including the LLM).

%% ========================================================================
\section{Main Results}
\label{sec:results}

\subsection{Oracle Headroom Exists}
\label{sec:oracle}

\begin{figure}[t]
\centering
\includegraphics[width=\columnwidth]{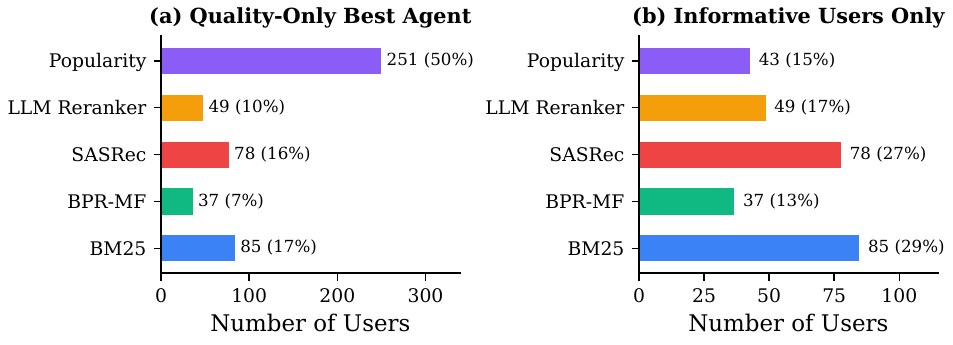}
\caption{Distribution of the quality-best agent per user. Panel~(a) includes all 500 users, including all-zero cases where deterministic tie handling creates a large zero-quality mass; panel~(b) focuses on the 292 informative users with $\geq$1 agent at NDCG$>$0. Evaluation uses the primary 5-agent pool with Qwen as the default LLM agent.}
\label{fig:oracle}
\Description{Two horizontal bar charts showing the distribution of the quality-best agent per user. The informative-user panel distributes best-agent labels across BM25, SASRec, LLM reranker, Popularity, and BPR-MF.}
\end{figure}

Before evaluating trainable methods, we verify the fundamental premise: \emph{no single agent is best for all informative requests}.
Fig.~\ref{fig:oracle} shows that among the 292 users where at least one agent achieves NDCG$>$0, the quality-best agent is distributed across all five agents: BM25 (29.1\%), SASRec (26.7\%), LLM Reranker (16.8\%), Popularity (14.7\%), and BPR-MF (12.7\%).
This confirms that the agent pool contains substantial complementary signal.

\subsection{Main Comparison}

Table~\ref{tab:main} presents the full comparison.
Single-agent, unsupervised fusion, and oracle rows are deterministic evaluations over the 500-user universe.
All trainable rows use 5-fold out-of-fold results averaged over three seeds.
For budgeted stackers, the name gives the validation LLM-budget cap; the LLM\% column gives the realised out-of-fold test ratio.
Table~\ref{tab:cost} reports the corresponding policy-level latency and token proxies for representative methods.

\begin{table}[t]
\centering
\caption{Main comparison on MovieLens-1M. Trainable rows use strict 5-fold out-of-fold evaluation (mean over 3 seeds; standard deviations shown where applicable). Budget rows select validation thresholds under LLM budget caps. \textbf{Bold} marks the best non-oracle quality or utility value; LLM\% is a cost axis, so lower usage is not treated as an unqualified winner.}
\label{tab:main}
\scriptsize
\resizebox{\columnwidth}{!}{%
\begin{tabular}{lccccc}
\toprule
\textbf{Method} & \textbf{HR@10$\uparrow$} & \textbf{NDCG@10$\uparrow$} & \textbf{MRR$\uparrow$} & \textbf{Utility$\uparrow$} & \textbf{LLM\%$\downarrow$} \\
\midrule
\multicolumn{6}{l}{\emph{Single agents}} \\
\quad Popularity      & 0.170 & 0.072 & 0.043 & 0.157 & 0\% \\
\quad BPR-MF          & 0.180 & 0.081 & 0.052 & 0.171 & 0\% \\
\quad SASRec          & 0.214 & 0.118 & 0.089 & 0.225 & 0\% \\
\quad BM25            & 0.254 & 0.114 & 0.072 & 0.240 & 0\% \\
\quad LLM Reranker    & 0.162 & 0.071 & 0.044 & 0.144 & 100\% \\
\midrule
\multicolumn{6}{l}{\emph{Fusion baselines}} \\
\quad RRF-cheap       & 0.200 & 0.092 & 0.060 & 0.192 & 0\% \\
\quad Borda-cheap     & 0.208 & 0.093 & 0.059 & 0.197 & 0\% \\
\quad ScoreAvg-cheap  & 0.232 & 0.101 & 0.062 & 0.216 & 0\% \\
\quad RRF             & 0.172 & 0.076 & 0.047 & 0.154 & 100\% \\
\quad Borda           & 0.178 & 0.078 & 0.049 & 0.160 & 100\% \\
\quad ScoreAvg        & 0.232 & 0.106 & 0.068 & 0.214 & 100\% \\
\midrule
\multicolumn{6}{l}{\emph{Selection baselines}} \\
\quad Random          & 0.214 & 0.103 & 0.071 & 0.209 & 18.4\% \\
\quad SingleBest      & 0.254 & 0.114 & 0.072 & 0.240 & 0\% \\
\quad ReqOnly OOF     & 0.219$\pm$0.013 & 0.105$\pm$0.004 & 0.071$\pm$0.001 & 0.214$\pm$0.010 & 6.5\% \\
\quad OneHot OOF      & 0.213$\pm$0.018 & 0.095$\pm$0.012 & 0.060$\pm$0.010 & 0.200$\pm$0.022 & 12.0\% \\
\midrule
\multicolumn{6}{l}{\emph{\routerec{}-Select hard selection (OOF)}} \\
\routerec{}-Select cheap & 0.223$\pm$0.008 & 0.102$\pm$0.003 & 0.067$\pm$0.003 & 0.214$\pm$0.006 & 0\% \\
\routerec{}-Select + gate & 0.215$\pm$0.016 & 0.099$\pm$0.010 & 0.064$\pm$0.008 & 0.205$\pm$0.018 & 10.4\% \\
\midrule
\multicolumn{6}{l}{\emph{Learned shortlist stacking (OOF)}} \\
\routerec{}-StackCheap & 0.254$\pm$0.007 & 0.123$\pm$0.004 & 0.084$\pm$0.003 & 0.250$\pm$0.007 & 0\% \\
\routerec{}-StackBudget@25 & 0.262$\pm$0.009 & 0.133$\pm$0.006 & 0.095$\pm$0.005 & 0.263$\pm$0.010 & 13.4\% \\
\routerec{}-StackBudget@50 & 0.279$\pm$0.011 & 0.145$\pm$0.012 & 0.105$\pm$0.012 & 0.281$\pm$0.017 & 38.5\% \\
\routerec{}-StackBudget@75 & 0.290$\pm$0.019 & 0.157$\pm$0.017 & 0.117$\pm$0.016 & 0.297$\pm$0.026 & 60.1\% \\
\routerec{}-StackGate & 0.295$\pm$0.022 & 0.161$\pm$0.019 & 0.122$\pm$0.018 & 0.303$\pm$0.030 & 70.2\% \\
\routerec{}-StackAll & \textbf{0.299$\pm$0.015} & \textbf{0.170$\pm$0.014} & \textbf{0.131$\pm$0.014} & \textbf{0.311$\pm$0.021} & 100\% \\
\midrule
\multicolumn{6}{l}{\emph{Quality upper bounds}} \\
\quad CheapOracle     & 0.508 & 0.260 & 0.185 & 0.514 & 0\% \\
\quad Oracle          & 0.584 & 0.299 & 0.214 & 0.590 & 9.8\% \\
\bottomrule
\end{tabular}%
}
\end{table}

\begin{table}[t]
\centering
\caption{Policy-level cost proxies for representative methods. Latency is summed cached wall-clock seconds over the invoked agent set; tokens are output tokens divided by 1000.}
\label{tab:cost}
\scriptsize
\begin{tabular}{lcccc}
\toprule
\textbf{Method} & \textbf{Utility} & \textbf{Latency(s)} & \textbf{Tok.} & \textbf{LLM\%} \\
\midrule
BM25 & 0.240 & 0.021 & 0.00 & 0.0\% \\
\routerec{}-Select cheap & 0.214 & 0.022 & 0.00 & 0.0\% \\
\routerec{}-Select + gate & 0.205 & 0.100 & 1.35 & 10.4\% \\
\routerec{}-StackCheap & 0.250 & 0.022 & 0.00 & 0.0\% \\
\routerec{}-StackBudget@25 & 0.263 & 0.123 & 1.74 & 13.4\% \\
\routerec{}-StackGate & 0.303 & 0.554 & 9.13 & 70.2\% \\
\routerec{}-StackAll & 0.311 & 0.780 & 13.00 & 100.0\% \\
\bottomrule
\end{tabular}
\end{table}

\begin{sloppypar}
Strict hard selection generalises weakly.
BM25 remains the best fixed agent, with HR@10 of 0.254.
\routerec{}-Select cheap-only obtains $0.223 \pm 0.008$, below BM25 but above SASRec.
The escalation gate obtains $0.215 \pm 0.016$ while using the LLM on 10.4\% of requests.
Learned shortlist aggregation changes the action space.
\routerec{}-StackCheap matches BM25 HR and has a higher NDCG point estimate.
With a 25\% validation budget, the budgeted stacker reaches $0.262 \pm 0.009$ HR@10 at 13.4\% LLM usage.
\routerec{}-StackGate reaches $0.295 \pm 0.022$ HR@10 at 70.2\%.
Several findings stand out.
CheapOracle reaches HR@10$=$0.508 and the full quality oracle reaches 0.584, so the failure is not lack of per-instance variation.
The hard selector wins over BM25 on only 1.2--3.2\% of users, ties on 90.0--95.4\%, and loses on 3.4--6.8\%, which explains why strict hard selection does not beat the best fixed agent.
Unsupervised fusion is also weak: cheap-only fusion underperforms StackCheap, and all-agent fusion underperforms BM25 despite invoking the LLM on every request.
Learned aggregation is the exception because deployable rank/score features recover useful cross-agent signal, and budgeted thresholds expose a gradual cost--quality frontier.
\end{sloppypar}

Paired user-level bootstrap against BM25 confirms the distinction.
\routerec{}-StackCheap's HR difference is indistinguishable from zero.
\routerec{}-StackGate improves HR by $+0.041$ (95\% CI $[0.005,0.075]$) and NDCG by $+0.048$ (CI $[0.024,0.073]$).

\begin{figure}[t]
\centering
\includegraphics[width=0.78\columnwidth]{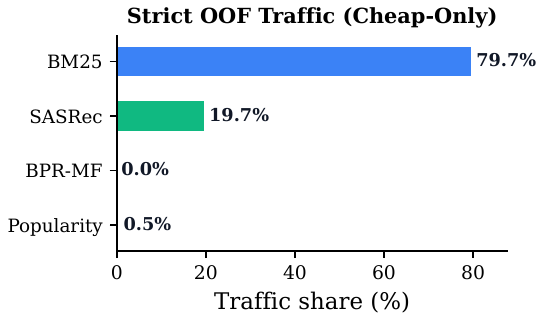}
\caption{\routerec{}-Select strict out-of-fold traffic distribution across cheap agents. The router mostly selects BM25 and SASRec; BPR-MF and Popularity receive zero or negligible traffic.}
\label{fig:traffic}
\Description{Horizontal bar chart showing RouteRec-Select traffic distribution across BM25, SASRec, BPR-MF, and Popularity.}
\end{figure}

Fig.~\ref{fig:traffic} shows the strict cheap-only \routerec{}-Select traffic distribution averaged over seeds: BM25 79.7\%, SASRec 19.7\%, Popularity 0.5\%, and BPR-MF 0.0\%.
The learned policy is therefore close to a BM25/SASRec switcher rather than a robust four-agent selector.

%% ========================================================================
\section{Ablations and Analysis}
\label{sec:ablations}

\subsection{Subgroup Analysis}

\begin{figure}[t]
\centering
\includegraphics[width=0.85\columnwidth]{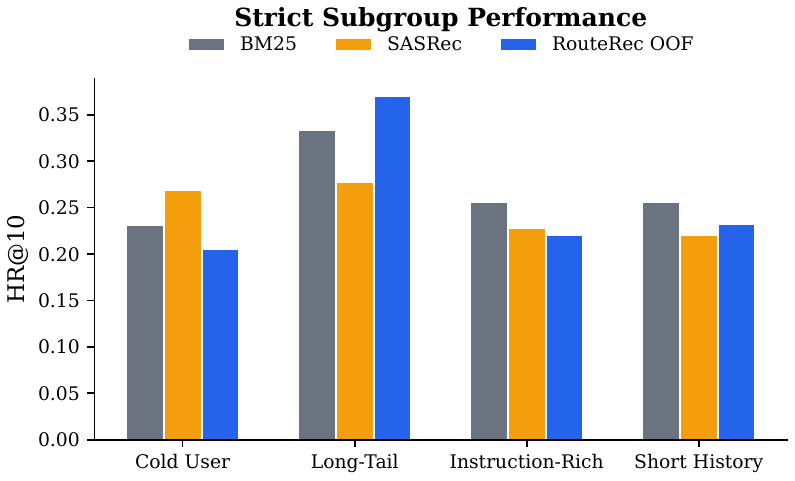}
\caption{Strict subgroup HR@10 comparison across BM25, SASRec, and \routerec{}-Select. The out-of-fold router remains conservative and only improves on the small long-tail slice.}
\label{fig:subgroup}
\Description{Grouped bar chart comparing BM25, SASRec, and RouteRec-Select HR at 10 across four subgroups: cold user, long-tail, instruction-rich, and short history. RouteRec-Select is strongest only on the small long-tail subgroup.}
\end{figure}

Table~\ref{tab:subgroup} and Fig.~\ref{fig:subgroup} show that \routerec{}-Select does not recover the subgroup gains suggested by the oracle analysis.

\begin{table}[t]
\centering
\caption{Strict subgroup comparison on HR@10 and NDCG@10. \routerec{}-Select uses cheap-only out-of-fold predictions averaged over three seeds.}
\label{tab:subgroup}
\small
\resizebox{\columnwidth}{!}{%
\begin{tabular}{l@{\hskip 4pt}cc@{\hskip 8pt}cc@{\hskip 8pt}cc}
\toprule
& \multicolumn{2}{c}{\textbf{BM25}} & \multicolumn{2}{c}{\textbf{SASRec}} & \multicolumn{2}{c}{\textbf{\routerec{}-Sel.}} \\
\cmidrule(lr){2-3} \cmidrule(lr){4-5} \cmidrule(lr){6-7}
\textbf{Subgroup} & HR & NDCG & HR & NDCG & HR & NDCG \\
\midrule
Cold (n=26)          & 0.231 & 0.110 & \textbf{0.269} & \textbf{0.190} & 0.205 & 0.103 \\
Long-tail (n=18)     & 0.333 & 0.187 & 0.278 & 0.158 & \textbf{0.370} & \textbf{0.222} \\
Instr.-rich (n=250)  & \textbf{0.256} & 0.117 & 0.228 & \textbf{0.123} & 0.220 & 0.101 \\
Short hist. (n=250)  & \textbf{0.256} & 0.115 & 0.220 & \textbf{0.123} & 0.232 & 0.106 \\
\bottomrule
\end{tabular}%
}
\end{table}

\routerec{}-Select remains weak on most subgroups under OOF evaluation.
The exception is the small long-tail slice, where the weighted hard selector beats BM25; this gain does not transfer to the full population.
This result does not establish reliable subgroup-specific routing from only 500 labelled users; the small slice sizes and sparse feedback leave substantial uncertainty about whether richer data could make the disagreement features useful.

\subsection{Retrain Ablation}
\label{sec:ablation}

To evaluate feature importance, we retrain each variant under the same strict out-of-fold protocol.
Table~\ref{tab:ablation} presents the results.

\begin{table}[t]
\centering
\caption{Strict out-of-fold ablations (means over three seeds). Feature-removal deltas are small at this sample size and do not support a reliable ranking of feature-family importance.}
\label{tab:ablation}
\small
\begin{tabular}{lcc}
\toprule
\textbf{Variant (retrained)} & \textbf{HR@10} & \textbf{NDCG@10} \\
\midrule
\textbf{Full \routerec{}-Select + gate} & 0.215 & 0.099 \\
\textbf{\routerec{}-Select cheap-only}  & 0.223 & 0.102 \\
\midrule
No request features    & 0.213 & 0.096 \\
No user features       & 0.208 & 0.093 \\
No candidate features  & 0.211 & 0.099 \\
No task features       & 0.221 & 0.100 \\
\midrule
No probes              & 0.219 & 0.105 \\
One-hot capability     & 0.213 & 0.095 \\
\bottomrule
\end{tabular}
\end{table}

\begin{figure}[t]
\centering
\includegraphics[width=0.85\columnwidth]{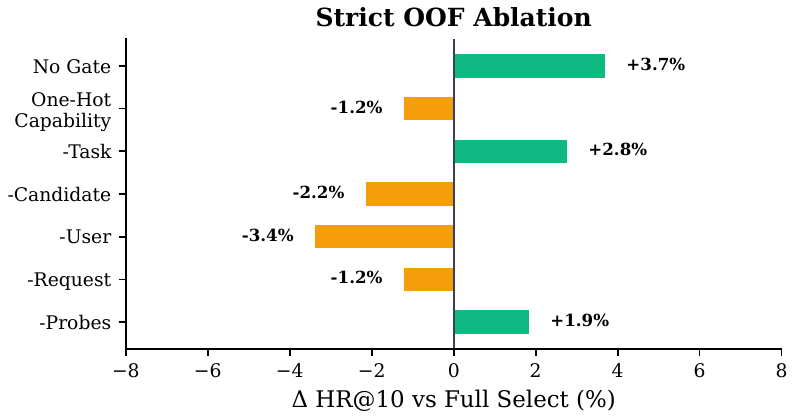}
\caption{Strict OOF ablation impact relative to full \routerec{}-Select. Differences are small and do not support a strong feature-importance claim.}
\label{fig:ablation}
\Description{Horizontal bar chart showing retrain ablation impact. All ablation variants remain close to full RouteRec-Select and below BM25.}
\end{figure}

The strict ablation pattern is flat relative to the observed run-to-run variation.
Removing probes, request features, or individual feature groups changes HR by only a few points around the full model's 0.215 baseline.
The best variant is cheap-only hard selection at HR $=0.223$, still below BM25.
These results do not establish that the feature families are uninformative; with only 500 users and one sparse leave-one-out label per user, the experiment has limited power to separate small feature effects from sampling variation.
Accordingly, we make no strong claim about probe--context synergy or feature importance under this feedback protocol.
Capability vectors and one-hot agent IDs also perform similarly, so a four-agent closed pool cannot meaningfully test capability generalisation.

\subsection{Escalation Gate Analysis}
\label{sec:gate}

Under the quality oracle (Section~\ref{sec:oracle}), the Qwen LLM reranker is the best agent for only 49/500 instances (9.8\%).
Under strict OOF evaluation, the learned gate does not reliably identify these instances.
At $\tau{=}0.1$, it reaches F1$=0.107{\pm}0.012$ while escalating 13.8\% of requests; at $\tau{=}0.2$, precision is $0.210{\pm}0.214$, recall is $0.027{\pm}0.031$, and F1 falls to $0.048{\pm}0.054$.
Validation-selected thresholds use the LLM on 10.4\% of requests on average, but HR drops from 0.223 for cheap-only routing to 0.215 with gating.
The gate is therefore not yet useful in this data regime.
Targeted analyses based on leakage-free disagreement features lead to the same conservative result: LLM escalation does not consistently improve out-of-fold routing, and the \routerec{}-Select+Gate curve remains below BM25 in the main population.
Thus instruction-rich or high-disagreement requests are not sufficient, in this sample, to identify LLM-worthy cases reliably; evaluating richer request semantics requires additional data.

\subsection{Stronger LLM under the Same Candidate Protocol}
\label{sec:grok}

To test whether a stronger backend changes the SASRec-seeded reranking picture, we replace the open-source Qwen-7B reranker with the \texttt{grok-4-1-fast-reasoning} backend~\cite{xai2026grok41}.
We use cached outputs available in our experimental logs and cite xAI's official Grok~4.1 Fast release page.
Using the \emph{identical} top-20 prompt and candidate-pool construction, Grok reaches HR@10 of 0.176, NDCG of 0.090, and MRR of 0.064.
This improves over Qwen (0.162/0.071/0.044) but remains \textbf{31\% below the best cheap agent} in HR, BM25 at 0.254.

In direct comparison, Grok outperforms Qwen on 59 of 500 instances (11.8\%); Qwen wins on 50 (10.0\%), and 391 are ties (78.2\%).
When both LLMs are in the candidate pool (HR-based comparison), Grok is the quality-best agent for 41 instances (8.2\%) and Qwen for 40 (8.0\%)---adding Grok increases the full quality oracle from 0.584 to 0.620.

Under the identical SASRec-seeded top-20 setup, replacing Qwen with Grok improves the LLM reranker but does not close the gap to BM25.
Together with the 36.4\% target-exposure rate, this shows that backend strength alone cannot overcome the candidate restriction; the experiment does not isolate how much of the remaining gap comes from model capability versus candidate construction.
Strict cheap-only \routerec{}-Select (HR $=0.223$) still exceeds both LLM rerankers but remains below BM25, so upgrading the LLM alone does not solve hard routing.

\subsection{Cost--Quality Curve}

\begin{figure}[t]
\centering
\includegraphics[width=0.85\columnwidth]{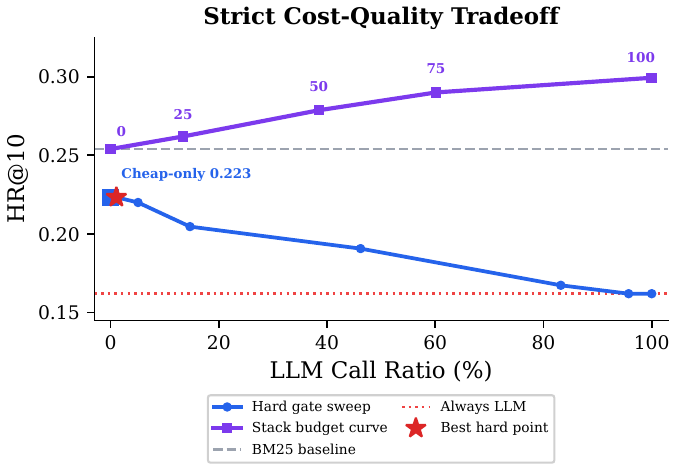}
\caption{Strict cost--quality curves. Hard-selection escalation does not improve quality, while budgeted learned stacking provides a smoother frontier between cheap-only stacking and all-agent stacking.}
\label{fig:cost_curve}
\Description{Line plot showing HR at 10 versus LLM call ratio under strict out-of-fold evaluation. It compares the hard gate sweep and the stack budget curve, with reference lines for SingleBest and Always-LLM.}
\end{figure}

Fig.~\ref{fig:cost_curve} shows recommendation quality as a function of the LLM call ratio.
The hard-selection curve is controlled by sweeping the gate threshold~$\tau$, while the stack curve uses validation-selected budget thresholds.
The strict hard-selection curve is mostly monotonic with respect to reducing LLM usage: always using the LLM gives HR $=0.162$, and the best hard-selection point occurs near no escalation (HR $\approx 0.223$).
Selective escalation therefore does not extract positive value for hard selection.
The stacker budget curve is different: \routerec{}-StackBudget@25 reaches HR $=0.262 \pm 0.009$ with 13.4\% LLM calls, \routerec{}-StackBudget@50 reaches $0.279 \pm 0.011$ with 38.5\% calls, \routerec{}-StackBudget@75 reaches $0.290 \pm 0.019$ with 60.1\% calls, and \routerec{}-StackAll reaches $0.299 \pm 0.015$.

\begin{figure}[t]
\centering
\includegraphics[width=\columnwidth]{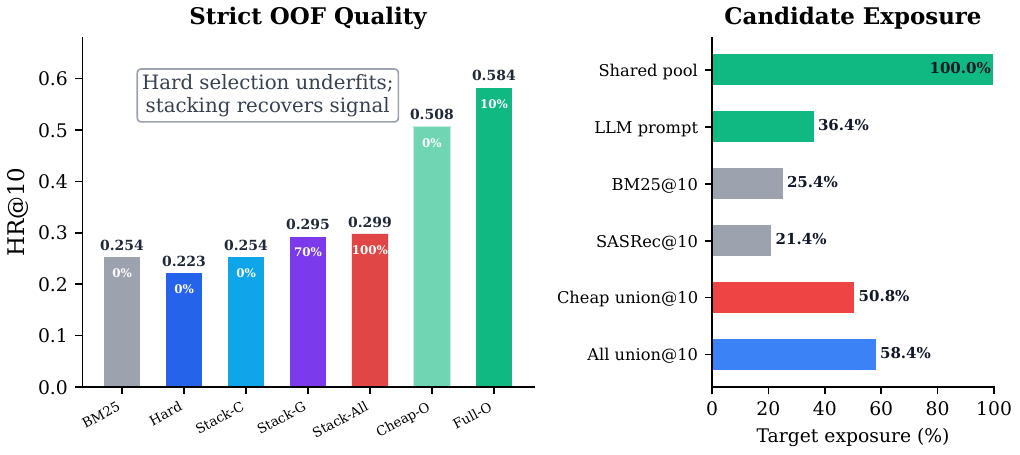}
\caption{Strict selection and aggregation versus quality-oracle headroom. Left: HR@10 comparison across BM25, hard-selection \routerec{}-Select, learned stacking variants, CheapOracle, and Oracle. Right: candidate exposure and oracle-coverage rates in the fixed-candidate benchmark.}
\label{fig:pareto}
\Description{Two-panel figure: left panel compares HR at 10 across BM25, RouteRec-Select, learned stacking variants, CheapOracle, and Oracle; right panel shows target exposure rates for the shared pool, LLM prompt, BM25, SASRec, cheap union, and all-agent union.}
\end{figure}

Fig.~\ref{fig:pareto} contrasts this learned behaviour with oracle headroom.
The large quality-oracle gap confirms that complementary candidate evidence exists.
Hard selection does not close this gap, whereas learned stacking recovers part of it.

%% ========================================================================
\section{Discussion}
\label{sec:discussion}

\paragraph{Oracle headroom vs.\ learnability.}
The central lesson is that oracle headroom is not enough.
The best agent varies substantially by request, but the labelled set is small and sparse: each user contributes one leave-one-out target, many agents tie at zero, and useful switching cases are rare.
The union of cheap top-10 lists contains the held-out item for 50.8\% of users, and the all-agent union reaches 58.4\%, so the missing signal exists in the candidate lists.
The hard selector fails because it must pick one whole list, not because alternative agents never contain the answer.
Under these conditions, a neural router can learn a plausible traffic distribution without learning a policy that improves HR.
This distinction is important for agent-search systems, where offline oracle analyses can make routing appear easier than it is.

\paragraph{When does LLM escalation help?}
In our fixed-candidate setup, Qwen-7B and Grok~4.1 Fast are quality-best for only $\sim$8--10\% of instances each.
Positive escalation examples are therefore sparse, and the strict learned gate does not find them reliably.
Replacing Qwen with Grok~4.1 Fast yields only a modest improvement (HR: 0.162$\to$0.176), still below BM25.
However, \routerec{}-StackAll's gains are consistent with the LLM shortlist contributing useful features to the combiner even though it is weak as a standalone answer.
This contrast highlights a structural difficulty in leave-one-out evaluation with SASRec-seeded candidates: useful LLM evidence need not arrive as the best complete list.

\paragraph{Selection vs.\ fusion.}
Unsupervised fusion baselines (RRF, Borda, score averaging) underperform the single-best agent, despite using all agents including the LLM.
Learned stacking is different: it calibrates per-agent ranks and scores on training users, so it can demote noisy list evidence.
The contrast shows that complementary agent outputs are not automatically useful: the learning problem and the unit of combination both matter.
In this setting, item-level reranking is a more effective action space than hard agent selection.

%% ========================================================================
\section{Conclusion}
\label{sec:conclusion}

We studied whether complementary outputs from heterogeneous recommender agents are better exploited by selecting one complete list or by aggregating item-level evidence.
Under strict out-of-fold evaluation, large oracle headroom does not translate into learnable hard routing: \routerec{}-Select remains below BM25, and selective LLM escalation does not improve it.
The same evidence becomes more useful at a finer decision granularity.
Cheap-only stacking matches BM25 in HR and has a higher NDCG point estimate, while all-agent stacking improves both metrics even though the LLM reranker is weak as a standalone agent.

The main lesson is methodological: routing is not solved.
Oracle complementarity, policy learnability, and decision granularity require separate evaluation.
In sparse offline settings, learned item-level aggregation can be a more promising use of heterogeneous outputs than choosing one agent's list wholesale.

\paragraph{Limitations.}
This is a controlled offline study on one dataset, 500 evaluation users, templated instructions, and a five-agent pool.
The SASRec-seeded fixed-candidate protocol restricts both LLM agents to top-20 reranking and exposes the held-out item to the LLM for only 36.4\% of requests.
It therefore does not test settings in which an LLM can influence candidate generation or use richer free-form intent, and the findings may not generalise to other datasets, domains, prompts, or retrieval pipelines.
Sparse leave-one-out supervision also limits the statistical power of the subgroup and feature ablations; their small deltas should not be read as evidence that the corresponding features are intrinsically useless.
Finally, scaling stacking beyond five agents may introduce redundant or conflicting lists, larger candidate unions, and additional calibration cost.
Our results consequently identify a failure mode of hard selection in this setting, not a universal ordering between routing and aggregation.

\paragraph{Future work.}
Future work should cover multiple domains, requests expressed in free form, alternative candidate-generation strategies, multi-item feedback, and larger open-ended agent pools.
Larger datasets could support context-dependent probe-reliability weighting and sparse or calibrated aggregation that remains tractable as the pool grows.
Online and real-time studies are also needed because user feedback and latency constraints can actively shape the selection--aggregation tradeoff.

\begingroup
\interlinepenalty=10000
\bibliographystyle{ACM-Reference-Format}
\bibliography{references}
\endgroup

\end{document}